\documentclass{article}
\usepackage{spconf,amsmath,graphicx}

\usepackage{latexsym}
\usepackage{amssymb}
\usepackage{amsmath}
\usepackage{amsthm}
\usepackage{booktabs}
\usepackage{enumitem}
\usepackage{graphicx}
\usepackage{svg}
\usepackage{float}
\usepackage{multirow}
\usepackage[acronym]{glossaries}
\makeglossaries
\setlist{nosep, leftmargin=14pt}

\usepackage{mwe} 

\title{Multimodal Oncology Agent for IDH1 Mutation Prediction in Low-Grade Glioma}

\name{Hafsa Akebli$^{1}$, Adam Shephard$^{2}$, Vincenzo Della Mea$^{1}$, and Nasir Rajpoot$^{2,3}$}
\address{
$^{1}$Department of Mathematics, Computer Science and Physics, University of Udine, Udine, Italy \\
$^{2}$Tissue Image Analytics Centre, Department of Computer Science, University of Warwick, Coventry, UK\\
$^{3}$Histofy Ltd, Coventry, UK
}

\begin{document}
\newacronym{WSI}{WSI}{Whole-Slide Image}
\newacronym{TCGA}{TCGA}{The Cancer Genome Atlas}
\newacronym{MLP}{MLP}{Multi-Layer Perceptron}
\newacronym{LGG}{LGG}{Low-Grade Glioma}
\newacronym{H&E}{H\&E}{Hematoxylin and Eosin}
\newacronym{AUROC}{AUROC}{Area Under the ROC Curve}
\newacronym{MOA}{MOA}{Multimodal Oncology Agent}
\maketitle
\begin{abstract}
Low-grade gliomas frequently present \textit{IDH1} mutations that define clinically distinct subgroups with specific prognostic and therapeutic implications. This work introduces a Multimodal Oncology Agent (MOA) integrating a histology tool based on the TITAN foundation model for \textit{IDH1} mutation prediction in low-grade glioma, combined with reasoning over structured clinical and genomic inputs through PubMed, Google Search, and OncoKB. MOA reports were quantitatively evaluated on 488 patients from the TCGA-LGG cohort against clinical and histology baselines. MOA without the histology tool outperformed the clinical baseline, achieving an F1-score of 0.826 compared to 0.798. When fused with histology features, MOA reached the highest performance with an F1-score of 0.912, exceeding both the histology baseline at 0.894 and the fused histology-clinical baseline at 0.897. These results demonstrate that the proposed agent captures complementary mutation-relevant information enriched through external biomedical sources, enabling accurate \textit{IDH1} mutation prediction.

\end{abstract}

\section{Introduction}
\label{sec:intro}
\textit{Isocitrate dehydrogenase~1} (\textit{IDH1}) mutations are highly prevalent in adult-type diffuse low-grade gliomas, occurring in approximately 50–81\% of WHO grade~II and~III tumors~\cite{IDH_Glioma}. These mutations identify a clinically important subgroup of gliomas characterized by slower progression, longer survival, and distinct therapeutic response profiles, making their accurate detection critical for diagnosis and prognostic assessment~\cite{IDH_Glioma,IDH_review}.

Most deep learning models for \textit{IDH1} mutation prediction are unimodal, relying on a single data source such as histopathology or radiomics and overlooking the broader clinical context~\cite{MRI_IDH1_DL}. Recent multimodal approaches integrating complementary sources of information, such as clinical, imaging, and histological data, have shown improved accuracy in \textit{IDH1} mutation prediction~\cite{DL_IDH1_Multimodal}.

Recently, agent-based architectures have been proposed to unify heterogeneous clinical data through large language model (LLM) reasoning, enabling interpretable and evidence-grounded decision-making in oncology. Such agents have been applied to early breast cancer detection using imaging and clinical data~\cite{breast_cancer_agent_nohistology}, and to hepatocellular carcinoma management integrating radiological and clinical information~\cite{HCC_agent}. The oncology agent introduced by Ferber~\textit{et~al.}~\cite{oncologyagent} integrates clinical, genomic, histological, and radiological data with external knowledge sources such as PubMed and OncoKB, demonstrating autonomous multimodal reasoning across gastrointestinal cancers.

In this work, we propose a \gls{MOA} for \textit{IDH1} mutation prediction in \gls{LGG}, using 488 patient cases from the TCGA-LGG cohort. We developed a histology tool based on the TITAN foundation model for \textit{IDH1} mutation prediction, which was integrated into the agent’s reasoning framework alongside structured clinical and genomic inputs. We quantitatively evaluated this agent by generating \gls{MOA} reports excluding the histology tool and comparing their predictive embeddings against clinical and histology baselines, as well as their fusion with histology features, enabling a systematic assessment of the \gls{MOA}'s multimodal reasoning performance. Our experiments demonstrate that the proposed agent effectively captures complementary, mutation-relevant information across modalities, enriched through external biomedical sources, for accurate \textit{IDH1} mutation prediction.

\section{Materials and Methods}
\subsection{The Dataset}
Experiments were conducted on the TCGA-LGG cohort, comprising diagnostic \glspl{WSI}, somatic mutation profiles, and clinical records, including demographic, diagnostic, and treatment data. Only patients with confirmed \textit{IDH1} mutation status were included, resulting in 488 cases (374 mutant and 114 wildtype). Each patient had one or more diagnostic \glspl{WSI} scanned at $40\times$ magnification; the slide with the largest tissue area was retained per patient. Radiology data were not included in this study.

\subsection{Histology Tool for \textit{IDH1} Status Prediction}
We built a histology tool for \textit{IDH1} mutation prediction and integrated it into the agent’s toolset. Diagnostic \glspl{WSI} were tiled into $512{\times}512$ patches at $20\times$ magnification, retaining regions with at least 90\% tissue content. Slide-level embeddings (768-dimensional) were extracted using TITAN~\cite{titan} from CONCH foundation model~\cite{CONCH} patch features and classified with a four-layer \gls{MLP}. Within the agent framework, the tool is dynamically invoked when a valid \gls{WSI} path or TITAN feature file is provided in the patient case, processing the input and returning an \textit{IDH1} mutation probability and binary prediction.

\subsection{MOA Configuration for \textit{IDH1} Status Prediction}
\begin{figure*}[ht!]
    \centering
    \includegraphics[width=\linewidth]{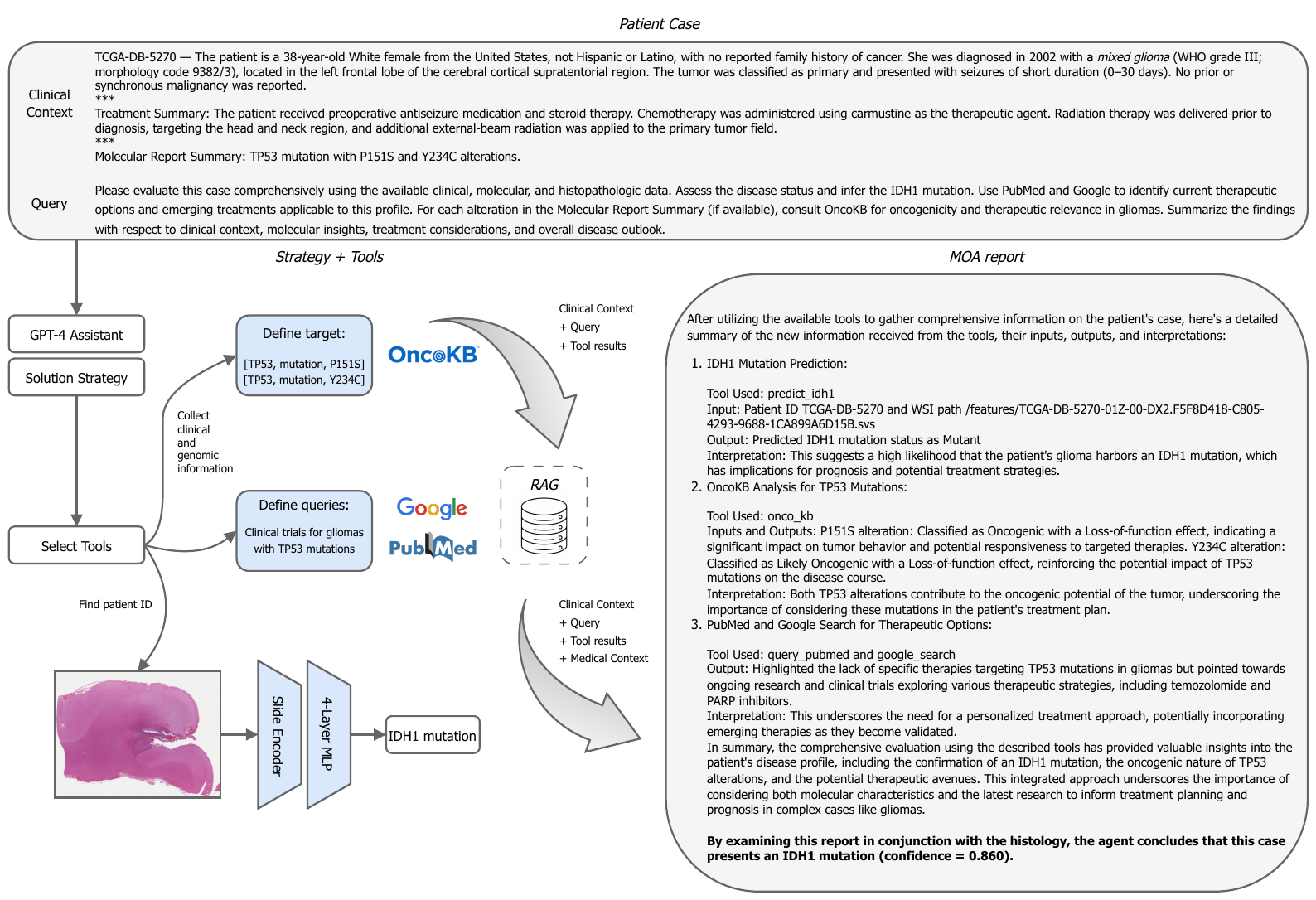}
    \caption{Demonstration of the MOA framework on a patient case. The GPT-4 Assistant autonomously selects tools based on the clinical context and query, retrieves evidence from OncoKB, PubMed, Google, and the histology tool for \textit{IDH1} mutation prediction, and integrates all outputs through RAG reasoning. The synthesized findings are compiled into an MOA report.}
    \label{fig:agent_framework}
\end{figure*}

Inspired by the oncology agent framework developed by Ferber~\textit{et~al.}~\cite{oncologyagent}, the proposed framework integrates multiple domain-specific tools, including PubMed and Google for literature and web evidence retrieval, and OncoKB~\cite{oncoKB} for curated knowledge on clinically actionable genomic alterations, enabling multimodal clinical reasoning. The MOA operates autonomously: given a textual patient context and query, MOA selects and invokes the relevant tools, integrates their outputs, and synthesizes them into a comprehensive report. For this study, the agent was configured for \textit{IDH1} mutation prediction in \gls{LGG}. The MedItron corpus of clinical guidelines~\cite{meditron} served as the primary knowledge base, filtered with glioma-specific keywords (e.g., glioma, oligodendroglioma) to retain 187 relevant documents, which were preprocessed and embedded into a local Chroma database for retrieval. Figure~\ref{fig:agent_framework} illustrates the configured \gls{MOA} workflow for \textit{IDH1} mutation prediction, showing how \gls{MOA} orchestrates tool use to generate the final report containing the \textit{IDH1} mutation status.

\subsection{MOA Reports for Quantitative Evaluation}
Unlike the original framework, which assessed agent generated reports qualitatively (e.g., for correctness or completeness), we quantitatively examined whether these reports encode measurable information related to \textit{IDH1} mutation status that can be detected through a predictive model.

\subsubsection{Report generation}
Reports were generated for all 488 patients by constructing structured patient cases containing demographic details (e.g., age, sex), diagnostic information (e.g., tumor class, histologic morphology), treatment records (e.g., treatment type, therapeutic procedure). Molecular summaries were derived from \gls{TCGA} somatic mutation profiles, filtered to retain genes frequently mutated in \gls{LGG} and annotated in OncoKB. Two genes, \textit{TP53} and \textit{CIC}, met these criteria, with validated oncogenic annotations available for 208 of the 488 patients. To prevent bias or label leakage, the histology tool, which outputs an \textit{IDH1} mutation probability, was disabled during report generation. Each constructed patient case, together with a fixed query applied uniformly across all patients, was then provided as input to the \gls{MOA}, which produced the final report by invoking the appropriate tools (PubMed, Google Search, and OncoKB).

\subsubsection{Report Embedding and Evaluation}
The 488 \gls{MOA} reports were cleaned and standardized before embedding. We used the gte-base-en-v1.5~\cite{gte} sentence transformer, a general-purpose long-context model supporting sequences up to 8{,}192 tokens. For comparison, the same encoder was applied to the clinical text of the 488 patients, derived from diagnosis, treatment, and demographic variables, while the same variables were also one-hot encoded as an additional baseline. The resulting embeddings were used as input to a four-layer \gls{MLP} trained for \textit{IDH1} mutation status prediction, quantifying how much mutation-related information is captured in the MOA reports and in the clinical data alone. Molecular summaries were excluded from the baselines due to missing values in 280 patients. 

\begin{figure}[t]
    \centering
    \includegraphics[width=\columnwidth]{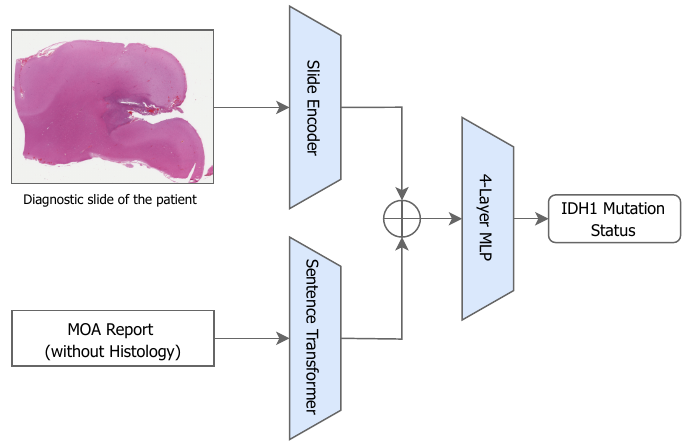}
    \caption{Integrated evaluation of the MOA for \textit{IDH1} mutation prediction. Slide embeddings are concatenated with MOA report embeddings (excluding the histology tool), obtained using a sentence transformer. The fused vector is classified by a 4-layer \gls{MLP} to predict the \textit{IDH1} mutation.}
    \label{fig:agent_evaluation}
\end{figure}

\subsection{Integrated Evaluation of the Complete Agent}
To assess the full oncology agent on \textit{IDH1} mutation prediction, the embeddings of MOA reports (excluding the histology tool) were concatenated with the corresponding slide embeddings extracted using TITAN. The fused representations were classified using the same four-layer \gls{MLP} as in previous experiments. This setup, illustrated in Fig.~\ref{fig:agent_evaluation}, quantifies the joint contribution of the agent’s textual reasoning and histological analysis to mutation-level prediction.

\section{Experiments and Results}
\subsection{Implementation details}
Experiments were conducted as an ablation study evaluating the \gls{MOA} with and without the histology tool. All experiments used five-fold stratified cross-validation and a four-layer \gls{MLP} trained with weighted cross-entropy loss. The Adam optimizer employed a learning rate of $1{\times}10^{-4}$ and weight decay of $1{\times}10^{-5}$. Text and histology embeddings were Z-score normalized. Models were trained with a batch size of 32 and evaluated with accuracy, F1-score, and the \gls{AUROC}.

\subsection{Agent Performance and Component Analysis}
All results are summarized in Table~\ref{tab:results}. The clinical text baseline, encoded with the sentence-transformer, achieved moderate performance (F1 = 0.789), while the one-hot encoding of structured clinical variables yielded slightly higher results (F1 = 0.798), showing that a simpler representation was more effective in this case. In contrast, the \gls{MOA} reports, encoded with the same sentence-transformer and without the use of the histology tool, achieved better overall performance (Accuracy = 0.802, F1 = 0.826, AUROC = 0.751) than both clinical baselines. This demonstrates that the agent enriches clinical information through contextual reasoning over patient data and external biomedical knowledge, capturing mutation-relevant patterns beyond explicit clinical features.

The histology tool achieved the strongest standalone performance (F1 = 0.894), confirming the high discriminative power of TITAN embeddings for \textit{IDH1} mutation prediction from \glspl{WSI}. When concatenated with clinical variables, the performance improved only slightly (F1 = 0.897), indicating that the histology features already capture key information related to \textit{IDH1} mutation. In contrast, fusing the \gls{MOA} report embeddings with histology features led to the highest overall performance (Accuracy = 0.915, F1 = 0.912, AUROC = 0.892). This demonstrates that the complete agent integrates complementary information from external biomedical knowledge with histological representations, leading to improved prediction of \textit{IDH1} mutation status.

\begin{table*}[t]
\centering
\caption{Performance of MOA variants with and without the histology tool, compared to clinical baselines and the standalone histology tool, for \textit{IDH1} mutation prediction on the TCGA-LGG cohort. Values are mean~$\pm$~std over 5 folds.}
\vspace{0.5em}
\renewcommand{\arraystretch}{1}
\begin{tabular}{lccccc}
\toprule
\textbf{Component Evaluated} & \textbf{Encoder} & \textbf{Model} & \textbf{Accuracy} & \textbf{F1} & \textbf{AUROC} \\
\midrule
\textbf{Clinical Text} & gte-base-en-v1.5 & MLP & $0.736{\pm}0.09$ & $0.789{\pm}0.05$ & $0.700{\pm}0.04$ \\
\textbf{Clinical Variables} & One-hot & MLP & $0.756{\pm}0.06$ & $0.798{\pm}0.02$ & $0.730{\pm}0.03$ \\
\textbf{MOA (without Histology)} & gte-base-en-v1.5 & MLP & $0.802{\pm}0.09$ & $0.826{\pm}0.05$ & $0.751{\pm}0.06$ \\
\midrule
\textbf{Histology Tool} & TITAN & MLP & $0.888{\pm}0.02$ & $0.894{\pm}0.02$ & $0.871{\pm}0.03$ \\
\textbf{Histology Tool + Clinical Variables} & One-hot + TITAN & MLP & $0.891{\pm}0.07$ & $0.897{\pm}0.04$ & $0.879{\pm}0.04$ \\
\textbf{MOA (with Histology)} & gte-base + TITAN & MLP & $\mathbf{0.915{\pm}0.02}$ & $\mathbf{0.912{\pm}0.02}$ & $\mathbf{0.892{\pm}0.04}$ \\
\bottomrule
\end{tabular}
\label{tab:results}
\end{table*}

\section{Conclusion and Future Directions}
In this paper, we proposed a multimodal oncology agent for \textit{IDH1} mutation prediction in \gls{LGG}. Unlike prior studies that evaluated agent reasoning qualitatively, we introduced a systematic protocol to measure the predictive content of \gls{MOA}-generated reports. The sentence-transformer used to encode the reports naturally handles missing information. The results show that text-based reasoning grounded in PubMed, Google Search, and OncoKB captures mutation-relevant information, while integration with a histology tool further enhances performance through complementary morphological features. 

Beyond predictive performance, an important aspect of the agent is its ability to handle missing data by adapting tool selection to the available inputs, demonstrating flexibility for real-world clinical application. Future directions include extending \gls{MOA} with radiology and pathology reports to enrich multimodal reasoning, and exploring open-weight language models for the agent’s reasoning and tool orchestration.

\section{Compliance with Ethical Standards}
This research study was conducted retrospectively using human subject data made available in open access by \gls{TCGA}. Ethical approval was not required, as confirmed by the license associated with the open-access data.

\section{Acknowledgements}
This work was carried out while the first author was visiting the TIA Centre. It was partially supported by the Next-Generation EU (PNRR: Piano Nazionale di Ripresa e Resilienza - Missione 4 Componente 2, D.M. 118/2023). Nasir Rajpoot is a co-founder and stakeholder in Histofy Ltd.

\begingroup
\renewcommand{\baselinestretch}{0.91}
\setlength{\itemsep}{0pt}
\small
\bibliographystyle{IEEEbib}
\bibliography{agent}

\begin{thebibliography}{10}

\bibitem{IDH_Glioma}
Ingo~K. Mellinghoff, Susan~M. Chang, and Kurt~A. Jaeckle,
\newblock ``Isocitrate dehydrogenase mutant grade ii and iii glial neoplasms,''
\newblock {\em Hematology/Oncology Clinics of North America}, vol. 36, no. 1, pp. 95--111, Feb. 2022.

\bibitem{IDH_review}
Julie~J Miller, L~Nicolas Gonzalez~Castro, and Samuel McBrayer,
\newblock ``Isocitrate dehydrogenase (idh) mutant gliomas: A society for neuro-oncology (sno) consensus review on diagnosis, management, and future directions,''
\newblock {\em Neuro-Oncology}, vol. 25, no. 1, pp. 4--25, 10 2022.

\bibitem{MRI_IDH1_DL}
Chandan~Ganesh Bangalore~Yogananda, Benjamin~C. Wagner, and Nghi C.~D. Truong,
\newblock ``Mri-based deep learning method for classification of idh mutation status,''
\newblock {\em Bioengineering}, vol. 10, no. 9, 2023.

\bibitem{DL_IDH1_Multimodal}
Riku Nakagaki, Shyam~Sundar Debsarkar, and Hiroharu Kawanaka,
\newblock ``Deep learning-based idh1 gene mutation prediction using histopathological imaging and clinical data,''
\newblock {\em Computers in Biology and Medicine}, vol. 179, pp. 108902, 2024.

\bibitem{breast_cancer_agent_nohistology}
Ilyass Emssaad, Fatima-Ezzahraa Ben-Bouazza, and Idriss Tafala,
\newblock ``Trustworthy multimodal ai agents for early breast cancer detection and clinical decision support,''
\newblock {\em Engineering Proceedings}, vol. 112, no. 1, 2025.

\bibitem{HCC_agent}
Liyang Wang, Fa~Tian, and Chengquan Li,
\newblock ``A multimodal llm-agent framework for personalized clinical decision-making in hepatocellular carcinoma,''
\newblock {\em Patterns}, p. 101364, 2025.

\bibitem{oncologyagent}
Dyke Ferber, Omar S.~M. El~Nahhas, and Georg Wölflein,
\newblock ``Development and validation of an autonomous artificial intelligence agent for clinical decision-making in oncology,''
\newblock {\em Nature Cancer}, vol. 6, no. 8, pp. 1337--1349, 2025.

\bibitem{titan}
Tong Ding, Sophia~J Wagner, and Andrew~H Song,
\newblock ``A multimodal whole-slide foundation model for pathology,''
\newblock {\em Nature Medicine}, 2025.

\bibitem{CONCH}
Ming~Y Lu, Bowen Chen, and Drew~FK Williamson,
\newblock ``A visual-language foundation model for computational pathology,''
\newblock {\em Nature Medicine}, vol. 30, pp. 863–874, 2024.

\bibitem{oncoKB}
Debyani Chakravarty, Jianjiong Gao, and Sarah Phillips,
\newblock ``Oncokb: A precision oncology knowledge base,''
\newblock {\em JCO Precision Oncology}, vol. 1, pp. 1--16, 2017.

\bibitem{meditron}
Zeming Chen, Alejandro~Hernández Cano, and Angelika Romanou,
\newblock ``Meditron-70b: Scaling medical pretraining for large language models,'' 2023.

\bibitem{gte}
Zehan Li, Xin Zhang, and Yanzhao Zhang,
\newblock ``Towards general text embeddings with multi-stage contrastive learning,'' 2023.

\end{thebibliography}
\endgroup

\end{document}